\documentclass[11pt]{article}

\usepackage[margin=1in]{geometry}
\usepackage{graphicx}
\usepackage{amsmath,amssymb}
\usepackage{booktabs}
\usepackage{hyperref}
\usepackage{xcolor}
\usepackage{caption}

\hypersetup{colorlinks=true, linkcolor=blue!55!black,
            citecolor=blue!55!black, urlcolor=blue!55!black}

\title{\bfseries Fractal KV-Cache Archives:\\
Lossless Symbolic Storage with In-Place Retrieval\\
for Long-Context LLM Inference}

\author{Vladimir Gusev\\[2pt]
\normalsize Independent Researcher\\[1pt]
\normalsize\texttt{vladimir@scriptum.ru}\\[1pt]
\normalsize\href{https://orcid.org/0009-0006-4236-9190}{ORCID:
0009-0006-4236-9190}}
\date{\today}

\newcommand{\ppl}[1]{\textbf{#1}}

\begin{document}
\maketitle

\begin{abstract}
The key--value (KV) cache dominates the memory cost of long-context
autoregressive inference, and a growing body of work compresses it through
quantization, eviction, or offloading. We study a complementary question:
once a position's KV state has been quantized to codebook indices, how
should the resulting \emph{symbol stream} be stored, and can the storage
layer do more than store? A family of contractive
iterated-map codes that serialize a symbol sequence into a sequence of
low-dimensional real vectors is revisited, and it is shown that they form a
natural \emph{archive} format for a quantized KV cache with the following
features. The method provides exactly the access pattern a growing cache requires. It is lossless, it runs in linear time, and supports $O(1)$
random access and $O(1)$ amortized append. A controlled study of the
quantizer feeding this archive is conducted on GPT-2 with 1024-token
contexts. Keeping a small exact window (4 attention
sinks $+$ 32 recent tokens) and archiving the rest, per-head residual
vector quantization reduces the archived cache by $36$--$54\times$
relative to an fp16 cache at a perplexity cost of $11$--$15\%$, and we
quantify a sharp key/value asymmetry---quantizing keys is roughly
$4\times$ more damaging than quantizing values, consistent with prior
low-bit KV work---and use it to allocate bits in a hybrid scheme. Finally, we show the archive is simultaneously a search index:
approximate substring queries execute directly on the stored vectors, and
matched context is decoded from the matched vector without ever
materializing the surrounding text. We further characterize the archive's
operating range: truncating the stored points yields a lossy regime whose
distortion we localize, and probability-weighting the contraction maps
recovers arithmetic coding, exposing a trade-off between rate efficiency,
random access, and memory.
We release all code; every number
reproduces from a single command on a laptop CPU.
\end{abstract}

\section{Introduction}

Serving a transformer language model over a long context is increasingly a
memory problem rather than a compute problem. To generate each new token
the model attends to cached key and value vectors for every previous
position in every layer; for GPT-2 this KV cache is $2 \times 12 \times 12
\times 64 = 18{,}432$ scalars per token, and for production-scale models it
reaches hundreds of kilobytes per token, so that for long documents the
cache dwarfs the model weights and becomes the binding constraint on batch
size and context length. The dominant responses are lossy: low-bit
quantization of the cached tensors, eviction of ``unimportant'' positions,
and offloading of cold cache to slower memory
\cite{sink,h2o,kvquant,kivi}.

This paper isolates a question that sits \emph{downstream} of the lossy
step and is usually left implicit. Suppose a position's KV state has
already been reduced to a handful of codebook indices by vector
quantization. Those indices are a symbol stream. How should the stream be
\emph{stored}, and--more interestingly--can the storage layer serve
retrieval and random access rather than being an inert blob? We show that a
lossless contractive iterated-map code answers both. Concretely, we test
three claims: (i) the code is a lossless, linear-time store with
sub-millisecond random access and append; (ii) with a small exact window,
per-head residual VQ archives the remaining cache at more than $30\times$
compression versus an fp16 cache for a bounded, measured perplexity cost;
and (iii) substring queries resolve directly on the stored vectors at full
recall. The success criterion for each is stated with its experiment.

\paragraph{The code.}
We use a contractive iterated-map code that reads a symbol sequence over an
alphabet of size $N$ and produces a trajectory of points in the plane: a
regular $N$-gon is fixed, one vertex per symbol, and each symbol $c_k$
maps the running point by $p_k = V(c_k) + r\,(p_{k-1} - V(c_k))$ for a
contraction ratio $r$. These $N$ contractions form an iterated function
system, and the set of points it can reach is self-similar: each symbol's
``cell'' is a scaled copy of the whole hull, which itself contains a scaled
copy for every next symbol, and so on at every depth. That self-similar
attractor --- the Sierpinski gasket in the three-symbol midpoint case --- is
the \emph{fractal} the title refers to; the code is simply a walk on it, one
step per symbol. Decoding inverts the map: the symbol is the vertex
whose contracted image (``cell'') contains the point, and
$p_{k-1} = (p_k - (1-r)V(c_k))/r$. Uniqueness requires the $N$ cells to be
disjoint, which bounds the ratio by a packing (``kissing'') constant
$r \le r_N$; the familiar midpoint code $r = \tfrac12$ is exactly the
boundary case for $N \le 4$ and must be tightened for larger alphabets.
Because inversion amplifies floating-point error by $1/r$ per step, one
IEEE-754 double reliably carries $\approx 44/\log_2(1/r)$ symbols, so a
practical codec stores one point per fixed-length span and reconstructs
each span by anchored backward search. The construction is classical in
the alignment-free sequence-analysis literature, where it is known as chaos
game representation and its generalization to arbitrary alphabets as
universal sequence maps \cite{jeffrey,almeida-usm}; our contribution is not
the code, nor vector quantization of the KV cache under a random-access
constraint (a concurrently active topic \cite{fibquant,retroinfer}), but
the use of this particular code as a KV archive that is at once a store, a
random-access decoder, and an in-place search index.

\paragraph{Contributions.}
\begin{enumerate}
  \item \textbf{A KV-archive format.} We frame a lossless contractive
  iterated-map code as the serialization layer for a quantized KV cache,
  and verify it is lossless, linear-time, and offers $O(1)$ random access
  to any past position and $O(1)$ amortized append--the operations a
  growing cache needs and that a general-purpose byte compressor does not
  provide (Section~\ref{sec:codec}).

  \item \textbf{A controlled quantizer study on long contexts.} With a
  small exact window and the remainder archived, we sweep pooled vs.
  per-head codebooks and residual depth on GPT-2 at 1024-token contexts.
  Per-head codebooks Pareto-dominate pooled ones at equal bit budget
  (Fig.~\ref{fig:pareto}), and we quantify a $\sim\!4\times$ key/value
  damage asymmetry (direction consistent with \cite{kivi}) that a
  bit-asymmetric hybrid exploits (Section~\ref{sec:vq}).

  \item \textbf{The archive is also an index.} Approximate substring
  retrieval runs \emph{directly} on the stored vectors--nearest-neighbor
  distance is a graded suffix-similarity--and matched context is decoded
  from the matched vector with no access to the surrounding text
  (Section~\ref{sec:retrieval}).
\end{enumerate}

All experiments use GPT-2 (124M) on a public-domain text and run in
minutes on a single CPU; code and scripts are released for full
reproduction at \url{https://github.com/eighteight/fractal-kv}.

\section{The codec as a KV archive}
\label{sec:codec}

We first establish that the code is a usable storage primitive
independent of any model. Table~\ref{tab:codec} reports a self-contained
benchmark of the codec on natural text: encoding is lossless across ASCII,
Unicode, and degenerate inputs; encoding time is linear in length
($0.68\,\mu$s/char, flat from $25$K to $400$K characters); random access to
an arbitrary character in a $10^6$-character document costs $311\,\mu$s
(decoding only the span that contains it); and appending a token costs
$175\,\mu$s amortized, re-walking only from the last stored anchor. These
are the primitives a KV archive requires: append as the cache grows, and
random access when a position is read back.

\begin{table}[t]
\centering
\caption{Codec primitives on natural text (single CPU). All round-trips
lossless.}
\label{tab:codec}
\begin{tabular}{lr}
\toprule
Operation & Cost \\
\midrule
Encode (linear, $25$K--$400$K chars) & $0.68\,\mu$s / char \\
Random access ($10^6$-char document)  & $311\,\mu$s / lookup \\
Append (token by token)               & $175\,\mu$s / token (amortized) \\
\bottomrule
\end{tabular}
\end{table}

\begin{figure}[t]
\centering
\includegraphics[width=0.86\linewidth]{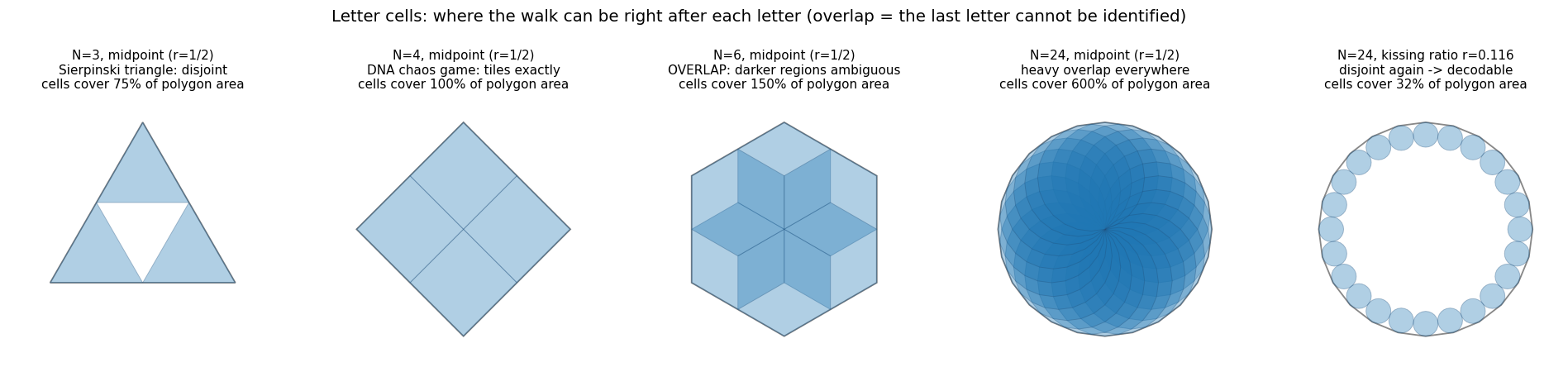}
\caption{Decodability of the code. Each symbol maps the polygon into a
contracted ``cell''; decoding the last symbol asks which cell a point
lies in, so cells must be disjoint. A half-scale (midpoint) cell has
$\tfrac14$ of the polygon's area, so $N$ cells need $N/4$ polygon-areas:
$N{=}3$ leaves gaps, $N{=}4$ tiles exactly, $N{>}4$ overlaps (ambiguous),
and shrinking $r$ to the kissing ratio restores disjointness for large
alphabets.}
\label{fig:cells}
\end{figure}

Figure~\ref{fig:cells} makes the alphabet/ratio constraint concrete: the
contraction ratio must adapt to alphabet size for the code to remain
uniquely decodable, which in turn sets the per-symbol precision budget and
hence the span length used by the codec.

\section{Quantizer study for the KV archive}
\label{sec:vq}

\paragraph{Setup.}
We evaluate on GPT-2 (124M) using \emph{The Time Machine} (Project
Gutenberg, public domain). Perplexity is measured token-by-token over a
$1024$-token passage. Codebooks are trained by $k$-means on KV vectors
collected from a \emph{disjoint} span of the same book. During evaluation
we keep an exact window--the first $4$ positions (attention sinks
\cite{sink}) and the most recent $32$--and, once a position ages out of
that window, overwrite its cached K and V with their nearest codebook
entries, so all later tokens attend to the quantized archive. This is the
regime of interest: at $1024$ tokens the exact window is $3.5\%$ of context
and the model attends almost entirely to archived, quantized memory.
Residual vector quantization (RVQ) of depth $s$ stores each vector as $s$
indices whose centroids sum to the reconstruction. Storage is reported as
bytes/token of the serialized index stream (fractal codec) and as
compression ratio against an fp16 cache ($36{,}864$ B/token); the exact
baseline perplexity is $35.97$.

\begin{figure}[t]
\centering
\includegraphics[width=0.78\linewidth]{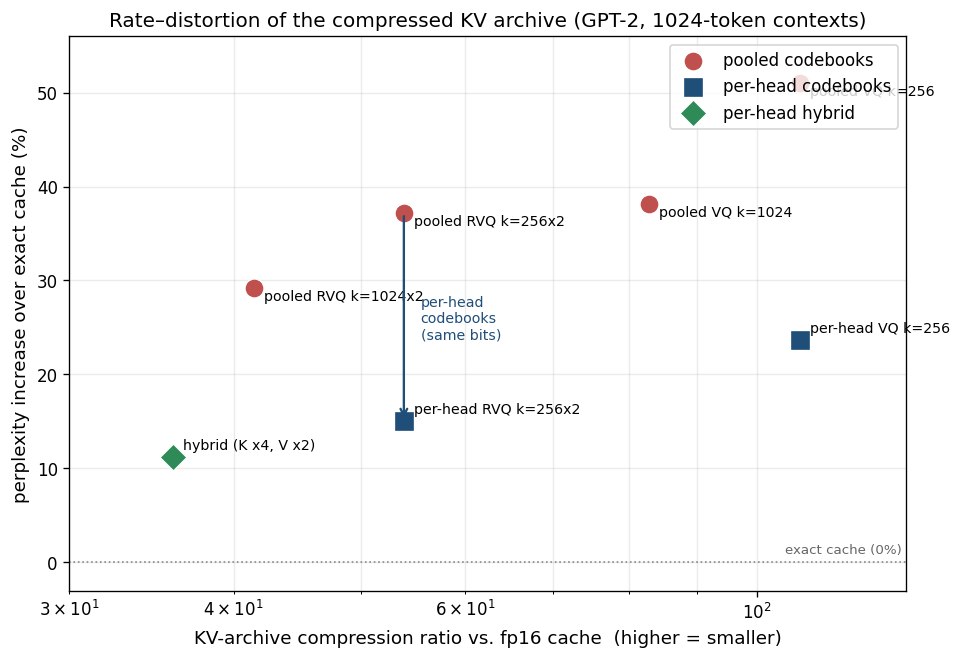}
\caption{Rate--distortion of the compressed KV archive. Per-head codebooks
(blue/green) Pareto-dominate pooled codebooks (red); the vertical arrow
marks the perplexity drop from pooling to per-head codebooks at an
identical bit budget. The bit-asymmetric hybrid attains the lowest
perplexity cost ($+11.2\%$) at $36\times$ compression.}
\label{fig:pareto}
\end{figure}

\paragraph{Per-head codebooks dominate.}
Table~\ref{tab:vq} and Figure~\ref{fig:pareto} give the main result.
Sharing one codebook across all heads of a layer (``pooled'') is
substantially worse than a codebook per head at \emph{equal bit budget}.
Single-stage per-head VQ costs $+23.7\%$ perplexity versus $+51.0\%$ for
pooled, and two-stage per-head RVQ costs $+15.0\%$ versus $+37.2\%$; in
both cases per-head codebooks move the entire rate--distortion curve down.
Each doubling of the index budget along the RVQ axis buys back a roughly
constant $8$--$9$ perplexity points, indicating a smooth operating curve
rather than a cliff. Per-head codebooks add a one-time $36$\,MB of
centroids ($288$ codebooks), trained in $108$\,s on CPU.

\begin{table}[t]
\centering
\caption{KV-archive quantization on GPT-2, $1024$-token contexts, $4$-sink
$+$ $32$-recent exact window. Perplexity increase is over the exact-cache
baseline ($35.97$); ``2nd half'' isolates the document half where archived
context dominates. Storage is the index stream via the fractal codec, and
compression is relative to an fp16 cache.}
\label{tab:vq}
\begin{tabular}{lccrr}
\toprule
Configuration & $\Delta$PPL & $\Delta$PPL (2nd half) &
  B/token & vs.\ fp16 \\
\midrule
Pooled VQ, $k{=}256$            & $+51.0\%$ & $+72.6\%$ & $342$  & $108\times$ \\
Pooled VQ, $k{=}1024$           & $+38.1\%$ & $+51.0\%$ & $445$  & $83\times$ \\
Pooled RVQ, $k{=}1024\times2$   & $+29.2\%$ & $+40.1\%$ & $889$  & $41\times$ \\
\midrule
Per-head VQ, $k{=}256$          & $+23.7\%$ & $+32.0\%$ & $342$  & $108\times$ \\
Per-head RVQ, $k{=}256\times2$  & $\ppl{+15.0\%}$ & $+22.4\%$ & $684$  & $54\times$ \\
\;\;\emph{values only (keys exact)} & $+4.0\%$  & $+6.7\%$  & -- & -- \\
\;\;\emph{keys only (values exact)} & $+14.5\%$ & $+20.7\%$ & -- & -- \\
\midrule
Per-head hybrid (K\,$\times4$, V\,$\times2$) & $\ppl{+11.2\%}$ & $+16.6\%$
  & $1025$ & $36\times$ \\
\bottomrule
\end{tabular}
\end{table}

\paragraph{Keys are $\sim\!4\times$ more fragile than values.}
Reusing the two-stage per-head codebooks, we quantize only values (keys
kept exact) and only keys (values kept exact). Quantizing values costs
$+4.0\%$ perplexity; quantizing keys costs $+14.5\%$--nearly the entire
$+15.0\%$ of quantizing both, and the two effects compose almost
independently ($1.145 \times 1.040 \approx 1.19$). The asymmetry has a
clear mechanism: values enter attention through a weighted average, which
attenuates per-vector error, whereas keys enter through the query--key dot
product that decides \emph{which} memories are read, so key error corrupts
attention routing itself. The direction of this effect matches the
key/value distinction reported by KIVI \cite{kivi}; our contribution is to
express it as a perplexity budget and to convert that budget into a bit
allocation.

\paragraph{A bit-asymmetric hybrid.}
Allocating four RVQ stages to keys and two to values lowers the cost to
$\ppl{+11.2\%}$ perplexity at $1025$ B/token ($36\times$ vs.\ fp16),
with keys-only-$\times4$ at $+9.4\%$ confirming the key axis is where the
budget is best spent. This is the best operating point in our sweep and
the recommended default; the key-bit scaling law suggests further stages
would continue to trade $\sim\!5$ points of perplexity per doubling.

\paragraph{On the storage layer's role.}
We are explicit that the compression \emph{ratio} in Table~\ref{tab:vq} is
produced by the vector quantizer; the fractal codec is the serialization
of the resulting index stream, and on these high-entropy indices its size
is comparable to bit-packing and to a general-purpose byte compressor
(within a few percent either way). Its value is not a smaller blob but a
richer one: unlike a byte compressor, the archive supports $O(1)$ random
access to any archived position and $O(1)$ append, and--as we show
next--doubles as a retrieval index over its own contents.

\section{Lossy operation: trading precision for memory}
\label{sec:lossy}

The archive so far is lossless, but a KV cache does not require exact
recovery of distant context, so it is natural to ask what is bought by
spending fewer bits per stored point and what is lost in return.
Truncating the code point to $B$ bits yields such a lossy code, and moves
the relevant bound from the entropy $H$ to the rate--distortion function
$R(D)$. The useful observation is \emph{where} the distortion falls.
Because each successive symbol refines the interval, the least significant
bits of the point describe the symbols coded last. Truncation therefore
destroys the tail of the coded sequence and leaves a contiguous exact
prefix. Coding the sequence in reverse inverts this: the exact region
becomes a contiguous \emph{suffix}, and the distant past degrades first.

\begin{table}[t]
\centering
\caption{Lossy truncation of a 240-character code point (order-0 model).
Forward coding preserves a contiguous prefix; reversed coding preserves a
contiguous suffix, i.e.\ recent context is recovered exactly while the
distant past degrades.}
\label{tab:lossy}
\begin{tabular}{rrlrl}
\toprule
& \multicolumn{2}{c}{Forward} & \multicolumn{2}{c}{Reversed} \\
\cmidrule(lr){2-3}\cmidrule(lr){4-5}
Bits kept & errors & exact region & errors & exact region \\
\midrule
200  & 195 & $[0..38]$  & 196 & $[204..239]$ \\
400  & 158 & $[0..72]$  & 161 & $[170..239]$ \\
800  & 86  & $[0..151]$ & 83  & $[90..239]$  \\
1200 & 16  & $[0..223]$ & 17  & $[18..239]$  \\
\bottomrule
\end{tabular}
\end{table}

Table~\ref{tab:lossy} shows the effect: under reversed coding the exact
horizon extends monotonically with the retained budget (36, 70, 150 and 222
exact trailing characters).

\paragraph{The single-interval gradient does not survive segmentation.}
The preceding analysis assumes the whole sequence is carried by one point,
which finite precision forbids: the codec stores one point per fixed-length
span (Section~\ref{sec:codec}). Truncating every span's point uniformly
therefore damages the far end of \emph{each} span, producing one damaged
region per span rather than a single contiguous one. Measuring this at a
fixed budget of $1.45$ bits/char on a 960-character string, one interval
yields a single exact prefix, whereas four and sixteen intervals yield a
periodic pattern of alternating exact and damaged regions spanning the
entire sequence:
\begin{center}\ttfamily\footnotesize
\begin{tabular}{ll}
1 span  & \texttt{....................XXXXXXXXXXXXXXXXXXXXXXXXXXXXXXXXX}\\
4 spans & \texttt{....xXXXXXXXXXXXXXx....XXXXXXXXXXXXXX......XXXXXXXXXXX}\\
16 spans& \texttt{.XXXXxXXXxxXXX.XXXxxXXXxXXXX.XXXxxXXX.XXXX.XXXxxXXX.XX}\\
\end{tabular}
\end{center}
\noindent(\texttt{.} exact, \texttt{x}/\texttt{X} partially/mostly wrong.)
Because a practical codec is necessarily multi-span, uniform truncation
degrades recent context as much as distant context, which is the opposite
of what an attention workload requires. Coding order then governs only the
placement of damage \emph{within} each short span and is of little
consequence. A recency gradient must instead be produced by allocating
precision \emph{across} spans---holding recent spans exact and truncating
distant ones---which is precisely the explicit exact-window policy of
Section~\ref{sec:vq}. We therefore do not claim the lossy code supplies
that policy for free; the geometry contributes only the within-span
ordering, and the cross-span allocation remains a deliberate design choice.

Two further caveats bound the regime. First, a pure recency gradient is not
the correct importance ordering for attention: the earliest positions carry
disproportionate attention mass \cite{sink}, so a practical scheme must
still pin a small sink prefix, and our long-context measurements show that
degradation of distant context is what costs accuracy on retrieval-style
workloads. Second, the weighted and reversed variants are inherently
sequential to decode, so adopting them forfeits the random access and
in-place search that motivate the uniform archive. We report this regime as
a characterization of the design space rather than as a drop-in
replacement.

\section{The archive is also a retrieval index}
\label{sec:retrieval}

A property of the code makes the stored archive searchable without
decompression. Because each step contracts by $r$, a stored point is
dominated by the most recent symbols, older ones decaying geometrically;
two positions sharing an $s$-symbol suffix therefore have points within
$2r^{s}$ of each other. Nearest-neighbor distance in point space is thus a
\emph{graded} suffix similarity, and thresholding at $2r^{\ell}$ retrieves
all positions whose length-$\ell$ suffix matches a query--a property known
for the midpoint code in the sequence-analysis literature
\cite{almeida-pattern} and which we measure directly and exploit for
KV-archive retrieval.

\begin{figure}[t]
\centering
\includegraphics[width=\linewidth]{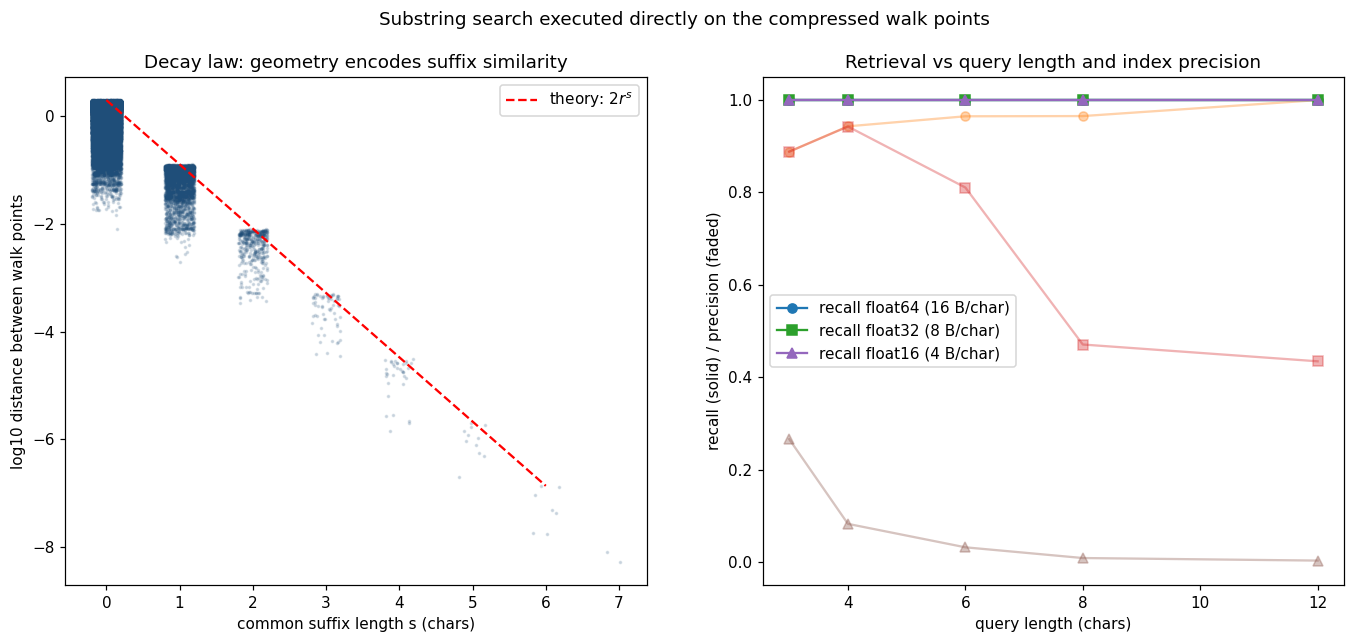}
\caption{Retrieval directly on the compressed archive. \textbf{Left:}
distance between stored points versus common-suffix length follows the
$2r^{s}$ decay law over seven decades. \textbf{Right:} retrieval recall
(solid) is $1.00$ at all query lengths; precision (faded) is governed by
index numerical precision relative to $r^{\ell}$, so a double-precision
index stays near-perfect while lower precision serves as a coarse
pre-filter.}
\label{fig:retrieval}
\end{figure}

Figure~\ref{fig:retrieval} confirms the decay law across seven orders of
magnitude on a $100$K-character corpus and reports retrieval quality. We
encode a query from the origin and threshold squared distance over the
stored points. Recall is $1.00$ at every query length by
construction--a true match \emph{must} fall inside the ball--while
precision is set by the index's numerical precision relative to $r^{\ell}$:
with a double-precision index ($16$\,B/char) precision is
$0.89$--$1.00$; with single precision ($8$\,B/char) recall is still
$1.00$ and the modest false-positive set is removable by decode
verification, which itself runs inside the representation. Because the
index lives in the plane rather than in the model's hidden dimension,
brute-force nearest-neighbor over the two-dimensional points costs only
$\approx\!0.9$\,ms/query over $100$K positions (and is trivially
accelerated by any spatial index)---this is a $2$D search, not
high-dimensional approximate nearest neighbor, and should not be compared
to embedding-space retrieval benchmarks. Crucially, the end-to-end loop
touches no external text: a query returns a matched point, and the
surrounding context is decoded \emph{backward from that point} through
stored anchors, reproducing the ground-truth context character for
character. In an inference setting this is exactly the operation a
retrieval-augmented decoder needs from its own past---locate archived
positions whose recent context matches a probe, and rehydrate only those,
without scanning or decompressing the rest of the cache.

\section{Relation to entropy coding}
\label{sec:limit}

The archive of Section~\ref{sec:codec} is lossless but makes no use of
source statistics: every symbol consumes $\log_2 N$ bits regardless of its
probability. This section places that design choice against the entropy bound, to be
explicit about what the choice costs and why it is nonetheless the right
one for an addressable archive.

\paragraph{The code is a degenerate arithmetic coder.}
Writing the contraction ratio per symbol rather than globally, let symbol
$c$ occupy a sub-interval of width $p(c)$ with cumulative offset $F(c)$.
Encoding maintains $[\mathit{lo}, \mathit{hi})$ and applies
\begin{equation}
  \mathit{lo}' = \mathit{lo} + w\,F(c), \qquad
  \mathit{hi}' = \mathit{lo} + w\,(F(c) + p(c)), \qquad
  w = \mathit{hi} - \mathit{lo},
\end{equation}
which is an iterated function system whose maps are sized by the model.
The code length of a string is then $-\sum_k \log_2 p(c_k)$, the model's
cross-entropy. Setting $p(c) = 1/N$ for all $c$ recovers the uniform
construction used above; sizing the maps by a source model recovers
arithmetic coding. The uniform code is therefore the special case of an
entropy coder under a uniform (i.e.\ worst-case) source model, and its rate
is bounded below by $\log_2 N$ rather than by $H(X)$. We note this
equivalence to situate the construction, not as a new result: chaos game
representation is a standard IFS on symbolic sequences
\cite{jeffrey,almeida-usm}, and interval subdivision by probability is the
textbook definition of arithmetic coding.

\paragraph{The cost, measured.}
Table~\ref{tab:limit} reports adaptive (prequential, add-one smoothed)
$n$-gram code lengths on 60{,}000 characters of non-repetitive English
prose, alongside general-purpose compressors. The uniform code spends
$6.13$ bits/char; a trivial order-2 model spends $3.58$, so the uniform
parameterization forgoes roughly $2.55$ bits/char, a factor of $1.71$.
Against Shannon's $1$--$1.5$ bits/char estimate for English the achievable
gap is larger still. We also observe the familiar sparsity effect: orders
above two degrade under add-one smoothing, as high-order contexts are
mostly unseen at this corpus size.

\begin{table}[t]
\centering
\caption{Rate on 60K characters of non-repetitive English prose
(alphabet $N = 70$). The uniform (model-free) code is the row the archive
of Section~\ref{sec:codec} implements.}
\label{tab:limit}
\begin{tabular}{lr}
\toprule
Method & bits/char \\
\midrule
Uniform code, $\log_2 N$ (this archive) & 6.129 \\
Probability-weighted, adaptive order-0  & 4.391 \\
Probability-weighted, adaptive order-1  & 3.611 \\
Probability-weighted, adaptive order-2  & \textbf{3.577} \\
zlib $-9$                               & 3.161 \\
lzma                                    & 2.970 \\
\midrule
Shannon estimate for English (reference) & $\sim$1--1.5 \\
\bottomrule
\end{tabular}
\end{table}

The conclusion we draw is deliberately narrow. As a \emph{compressor} on
structured text the uniform code is not competitive, and no tuning closes
that gap, because the gap \emph{is} the absent source model. What the
uniform parameterization buys instead is the structural property the
weighted version destroys: a fixed, position-addressable geometry, hence
$O(1)$ random access, $O(1)$ append, and in-place suffix search. Arithmetic
coding attains the entropy bound but emits a single opaque number that must
be decoded sequentially from the start. The design space is therefore a
three-way trade-off---rate efficiency, random access, and memory---and the
two parameterizations sit at opposite corners of it.

\section{Related work}

KV-cache compression spans low-bit quantization \cite{kvquant,kivi},
eviction of low-importance positions \cite{h2o}, and the observation that a
few initial ``sink'' tokens carry disproportionate attention mass
\cite{sink}; our exact-window scheme is a minimal combination of a sink
prefix and a recency window. The key/value asymmetry we measure is not new:
KIVI \cite{kivi} already reports that keys and values have distinct
distributions and require different quantization treatment (keys
per-channel, values per-token); our contribution here is only to quantify
the asymmetry as a perplexity budget on long contexts and to use it to
allocate residual-VQ depth. Vector quantization of the KV cache under a
random-access constraint is likewise an active topic: the concurrent
FibQuant \cite{fibquant} is, like us, a fixed-rate vector quantizer for
random-access KV compression evaluated on GPT-2, tracing a $5$--$34\times$
memory--fidelity frontier, and vector storage engines for long-context
inference are emerging \cite{retroinfer}. Relative to FibQuant, which
optimizes the codebook geometry, our quantizer study is deliberately plain
(k-means / residual VQ) and our distinct claim is about the \emph{storage
representation}: the contractive iterated-map code---chaos game
representation \cite{jeffrey} and its arbitrary-alphabet generalization,
universal sequence maps \cite{almeida-usm}---is simultaneously a lossless,
randomly-accessible, appendable store \emph{and} an in-place suffix-search
index, the latter resting on a decay property established for that code
\cite{almeida-pattern}. To our knowledge that unification of storage and
in-representation retrieval has not been applied to the KV-archive setting;
it, rather than the quantizer or the code in isolation, is what this paper
contributes. We developed this representation independently of, and
contemporaneously with, FibQuant \cite{fibquant}; we became aware of it, and
of RetroInfer \cite{retroinfer}, only after the experiments reported here
were complete, and we cite both as related work rather than claiming
priority over the overlap.

\section{Limitations}

Our study is deliberately small-scale and the claims should be read
accordingly. All results are on a single $124$M-parameter model, a single
public-domain corpus, and a single $1024$-token context length, using
perplexity as the only quality metric; downstream-task accuracy and larger
models are untested, and larger models are empirically more tolerant of KV
quantization, so these numbers are plausibly conservative but not
validated as such. The compression comes from the vector quantizer, and
our per-head codebooks are trained per corpus rather than amortized across
data; a deployable system would need corpus-independent or online
codebooks. The retrieval demonstration is exact/approximate \emph{suffix}
matching over token strings, not semantic retrieval over embeddings--the
stored vectors are not shown to encode meaning, only sequence structure.
Finally, our serialization ties rather than beats byte-oriented coders on
raw size; its advantage is the access pattern, not the ratio. We also do
not benchmark against concurrent random-access KV quantizers such as
FibQuant \cite{fibquant}, whose codebook design is complementary to and
likely stronger than our plain k-means; a fair head-to-head, and combining
their quantizer with our retrieval-capable storage, is left to future work.

\section{Conclusion}

Treating the storage of a quantized KV cache as a first-class design
choice, we showed that a lossless contractive iterated-map code is a
practical \emph{archive} format: linear-time, randomly accessible,
appendable, and--uniquely--searchable in place. Around it we conducted a
controlled quantizer study on long contexts, finding that per-head
residual VQ Pareto-dominates pooled codebooks, that key quantization is
about $4\times$ more damaging than value quantization, and that a
bit-asymmetric hybrid compresses the archived cache $36\times$ versus fp16
at $+11\%$ perplexity on GPT-2. The retrieval result points to the most
interesting direction: an inference-time memory that is stored, decoded,
and searched in one representation, so that a query can locate the archived
positions whose context resembles it without rehydrating the cache. We
release all code and scripts at
\url{https://github.com/eighteight/fractal-kv}; every figure and table
reproduces from a single command on a laptop CPU.

\end{document}